\documentclass[runningheads]{llncs}

\usepackage{eccv}

\usepackage{eccvabbrv}

\usepackage{graphicx}
\usepackage{booktabs}

\usepackage[accsupp]{axessibility}  

\usepackage{hyperref}

\usepackage{orcidlink}

\usepackage{booktabs}
\usepackage{graphicx}
\usepackage{pifont}

\newcommand{\cmark}{\ding{51}}
\newcommand{\xmark}{\ding{55}}

\usepackage{multirow}
\usepackage{makecell}

\newcommand{\modelcell}[2]{%
\multirow{#1}{*}{\makecell[c]{#2}}%
}

\begin{document}

\title{Seeing Before Answering: Training-Free Visual Layer Profiling for Vision-Language Models} 

\titlerunning{Training-Free Visual Layer Profiling for VLMs}

\author{Ruchen Liu\inst{1}\fnmsep\thanks{These authors contribute equally to this work.}\orcidlink{0009-0005-0465-1729} \and
Yi Yang\inst{1}\fnmsep$^{\star}$\orcidlink{0009-0001-1362-2269} \and
Yiming Xu\inst{1}\fnmsep$^{\star}$\orcidlink{0009-0004-6303-7801} \and
Michael Ying Yang\inst{2}\orcidlink{0000-0002-0649-9987} \and \\
Monika Sester\inst{1}\orcidlink{0000-0002-6656-8809} \and
Bodo Rosenhahn\inst{1}\fnmsep\thanks{Corresponding author.}\orcidlink{0000-0003-3861-1424}}

\authorrunning{R.~Liu et al.}

\institute{
Leibniz Universität Hannover, Germany \and
University of Bath, UK
}

\maketitle

\begin{abstract}

  LLaVA-style Vision-Language Models (VLMs) pass visual tokens from a fixed late layer of the vision backbone, typically the penultimate one, to the language model. 
  We first show that this hidden convention is fragile: across 2 VLMs and 7 image and video benchmarks, the default layer is sub-optimal in 13 of 14 model-task pairs, and the best layer shifts with both task and visual backbone. 
  Finding that layer by exhaustive layer-wise inference is prohibitively expensive, and no better fixed default exists.
  We therefore ask whether layer usefulness can instead be predicted from representation geometry.
  We study matrix-based entropy, introduced for unimodal layer analysis, which we compute over sample-level visual embeddings as Visual Dataset Entropy (VDE); and Gromov-Wasserstein (GW) distance, introduced for encoder-level VLM model selection, which we repurpose as a layer-wise visual--language alignment signal. 
  Transferring these to LLaVA-based models is not obvious a priori: the vision tower is frozen while the multimodal projector is trained, so we profile both sides of the projector.
  We find that VDE transfers, and GW does not. Computed from 100 unlabeled task samples without downstream inference, pre-projector VDE tracks layer-wise accuracy and its top-ranked layers cover the oracle best layer on every task for the SigLIP-based LLaVA-Video, while giving region-level guidance for the CLIP-based Video-LLaVA. 
  Post-projector profiles show that the projector reshapes visual geometry but does not erase the performance-relevant trend, leaving $\mathrm{VDE}_{\mathrm{pre}}$ the stronger signal. GW instead flattens after projection and is best read as an alignment diagnostic rather than a selector. VDE thus offers an interpretable, training-free policy that narrows the visual-layer search to a handful of candidates for limited downstream verification.
\end{abstract}

\section{Introduction}

Many Vision-Language Models (VLMs) use a single fixed layer of the vision backbone as the interface to the language model, and the widely-adopted LLaVA family~\cite{li2025llava,zhang2024llava,lin2024video} fixes this layer to a late depth, typically the penultimate one. This design is simple and widely adopted, but it makes the visual-layer choice a hidden architectural assumption: the same representation depth is expected to support different tasks, domains, and input modalities.

Layer-wise evaluation reveals that this assumption is fragile. Downstream performance varies systematically across visual depth, and the default layer is not always optimal, as shown in Table~\ref{tab:layer_comparison}. This effect appears across both image and video benchmarks, indicating that visual-layer choice is not merely an implementation detail but a meaningful factor in VLM behavior. Exhaustive layer-wise inference~\cite{chen-etal-2025-multimodal,liu2025visual,skean2025layer} can identify the best layer, but it requires repeated full evaluations and becomes costly for LLaVA-style large video VLMs.

\begin{table}[t]
\centering
\caption{Preliminary finding: the vision backbone's default layer is often sub-optimal.}
\label{tab:layer_comparison}
\resizebox{0.85\textwidth}{!}{
\begin{tabular}{lccc|ccc}
\toprule
\multirow{3}{*}{Task} &
\multicolumn{3}{c|}{LLaVA-Video~\cite{zhang2024llava} / SigLIP~\cite{Zhai_2023_ICCV}} &
\multicolumn{3}{c}{Video-LLaVA~\cite{lin2024video} / CLIP~\cite{radford2021learning}} \\
&
\multicolumn{3}{c|}{Default Layer = 26} &
\multicolumn{3}{c}{Default Layer = 23} \\
&
Default Acc. & Best Acc. & Best Layer &
Default Acc. & Best Acc. & Best Layer \\
\midrule
ScienceQA~\cite{lu2022learn}      & 87.62 & 88.90 & 27 & 59.63 & 59.63 & 23 \\
POPE-Adv.~\cite{li2023evaluating}      & 86.60 & 87.00 & 27 & 79.80 & 80.53 & 22 \\
CV-Bench 2D~\cite{tong2024cambrian}    & 68.50 & 70.20 & 27 & 49.86 & 53.20 & 20 \\
CV-Bench 3D~\cite{tong2024cambrian}    & 82.00 & 83.17 & 27 & 57.33 & 58.83 & 21 \\
MMStar~\cite{chen2024we}         & 57.20 & 58.07 & 27 & 32.40 & 33.00 & 21 \\
HD-EPIC Action~\cite{perrett2025hd} & 59.50 & 61.00 & 25 & 32.00 & 33.00 & 22 \\
HD-EPIC Gaze~\cite{perrett2025hd}   & 56.00 & 57.70 & 23 & 36.40 & 37.00 & 21 \\
\bottomrule
\end{tabular}
}
\vspace{-3mm}
\end{table}

This paper studies visual-layer choice as an interpretability problem. Rather than treating the best layer as a black-box empirical outcome, the analysis focuses on whether layer-wise representation geometry explains which visual layers are useful. The key finding is that sample-level visual diversity is strongly associated with downstream performance. Layers with richer and more discriminative sample geometry tend to form high-performing regions, and this signal is observable before answer generation.
The analysis is instantiated with Visual Dataset Entropy (VDE)~\cite{skean2025layer}, a matrix-entropy-based representation probe. Given a small set of task samples, VDE measures the diversity of sample-level visual embeddings at each candidate layer. VDE is computed both before and after the multimodal projector. The pre-projector profile reflects the intrinsic geometry of the vision backbone, while the post-projector profile captures how this geometry is reshaped at the vision-language interface.

Cross-modal structure is further examined with Gromov-Wasserstein (GW)~\cite{Li_2026_CVPR} distance to compare visual and language-side sample geometries. GW provides a coordinate-free diagnostic of visual-language compatibility across layers. In the experiments, GW reveals alignment trends across depth, while VDE provides the more reliable practical signal for layer profiling.

Experiments on LLaVA-Video~\cite{zhang2024llava} with a SigLIP~\cite{Zhai_2023_ICCV} vision tower and Video-LLaVA~\cite{lin2024video} with a CLIP~\cite{radford2021learning} vision tower show that high-VDE regions largely overlap with strong downstream layers across diverse image and video benchmarks. For the SigLIP-based model, VDE retrieves the best-performing layer within a small candidate set across all evaluated tasks. For the CLIP-based model, VDE is less exact as a strict Top-3 selector, but still identifies compact high-performing regions and filters out clearly weak layers. These results support entropy-based profiling as a simple, training-free policy for narrowing visual-layer search and as an interpretable tool for understanding VLM visual representations.

The main findings and contributions are summarized as follows:
\begin{itemize}
    \item Fixed late-layer extraction is shown to be a fragile convention in LLaVA-style VLMs: downstream performance changes systematically across visual depth, and the default layer is not always optimal.

    \item Visual Dataset Entropy is validated as a simple training-free policy for narrowing visual-layer search, using only a small set of task samples and no answer generation.

    \item The multimodal projector is shown to reshape layer-wise representation geometry, while Gromov-Wasserstein distance provides diagnostic evidence for visual-language alignment across depth.
\end{itemize}

\section{Related Work}

\noindent\textbf{Layer-wise Representation Analysis in Unimodal Models.}
Layer depth strongly shapes what a network encodes.
In vision transformers, representations evolve from local spatial structure in early layers to object-level and semantic abstractions in deeper ones~\cite{caron2021emerging,raghu2021vit,dosovitskiy2021vit}.
Parallel studies in language models, using linear probes~\cite{alain2016understanding} and intrinsic-geometry analyses~\cite{valeriani2023geometry,cheng2025emergence}, similarly find that intermediate rather than final layers often carry the most transferable or abstract information.
To characterize such structure without task labels, information-theoretic and spectral probes have been used, including representation rank, effective rank, and matrix-based entropy~\cite{wei2024diff,skean2023dime,roy2007effective}.
Matrix-based R\'enyi entropy in particular estimates information quantities directly from representation matrices and has been used to analyze information flow and layer-wise dynamics in deep networks~\cite{skean2023dime,wickstrom2019information,yu2020understanding}, revealing structured representation dynamics across both language and vision models~\cite{skean2025layer}.
These works establish entropy as an analytical tool for understanding internal representations, but remain confined to unimodal settings.

\noindent\textbf{Visual-Layer Selection and Cross-Modal Alignment in VLMs.}
Modern vision-lan\-guage models (VLMs) connect a vision tower such as CLIP or SigLIP to a large language model through a multimodal projector~\cite{radford2021learning,Zhai_2023_ICCV,zhang2024llava,lin2024video}, and typically feed the language model visual tokens from a single fixed layer; the widely-used LLaVA family fixes this to a late layer, commonly the penultimate one~\cite{zhang2024llava,lin2024video,li2025llava,marafioti2025smolvlm}.
This convention provides a stable interface for instruction tuning but decouples visual-layer choice from task requirements.
Recent work questions this default, showing that intermediate or shallower visual layers can be more effective for certain tasks~\cite{chen-etal-2025-multimodal} and motivating methods that fuse features from multiple visual layers~\cite{lin2025multi,cao2024mmfuser}.
A related line studies the projector itself, showing that its design affects cross-modal alignment and instruction tuning~\cite{liu2024improved,liu2023visual}; we instead ask how it reshapes layer-wise visual geometry at inference time.
Since visual and language features live in different spaces, direct comparison is difficult. 
Interpretability work carried out by Liu \etal traces how visual information propagates inside the language model, finding that its layers augment but can also degrade the visual signal received from the encoder~\cite{liu2025visual}. 
Gromov-Wasserstein (GW) distance quantitatively compares the relational geometry within each space~\cite{memoli2011gromov,peyre2016gromov} and has recently been applied to VLM model selection~\cite{Li_2026_CVPR}.
We adopt GW as one alignment probe and find it to be a useful diagnostic across visual layers, complementing entropy-based profiling.

\noindent\textbf{Summary.}
While layer-wise analysis is well developed for unimodal models, whether its conclusions transfer to VLM visual layers is underexplored, and few existing methods select the best visual layer without running full downstream inference.
We bridge this gap by extending representation profiling to VLMs with a training-free pipeline that predicts layer usefulness from sample-level visual geometry and presents it as a practical tool.

\begin{figure*}[t]
    \centering
    \includegraphics[width=\textwidth]{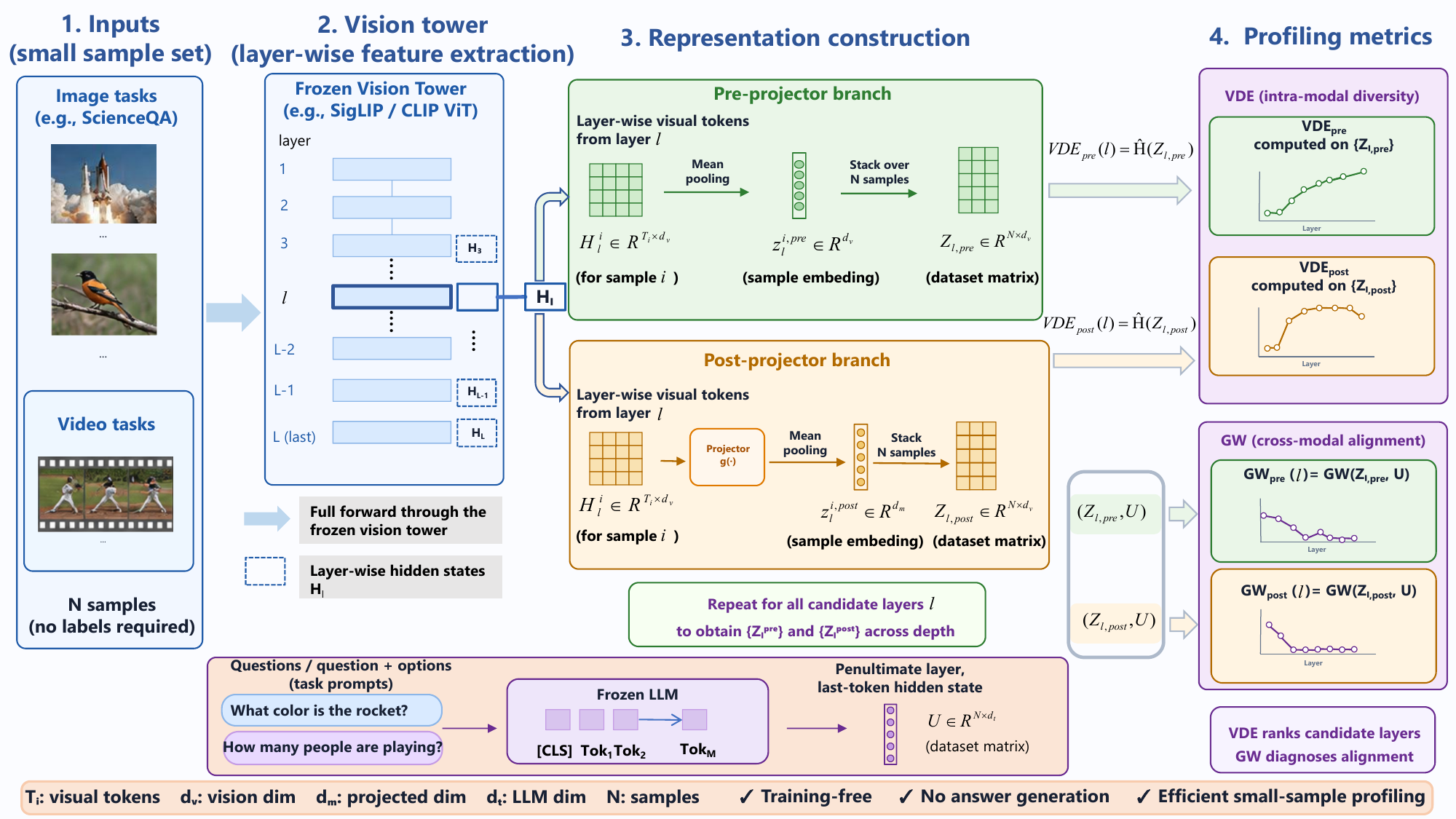}
    \caption{
    Overview of the proposed visual-layer profiling framework. Two training-free geometric signals are computed across candidate layers: VDE, measured from pre- and post-projector visual representations, and GW, measuring visual-language structural alignment. Both are profiled against downstream layer usefulness to assess how well each predicts strong visual layers.
    }
    \label{fig:framework}
    \vspace{-2mm}
\end{figure*}

\section{Method}
\label{sec:method}

We ask whether a small set of task samples can reveal which layer of a VLM vision tower yields the most effective visual representation, without answer generation, gradient updates, or exhaustive downstream inference. Figure~\ref{fig:framework} provides an overview of our visual-layer profiling framework. We focus on two training-free geometric signals from prior representation-analysis research: visual dataset entropy (VDE)~\cite{skean2025layer}, a matrix-based entropy; and Gromov-Wasserstein (GW) distance between visual and language features~\cite{peyre2016gromov,Li_2026_CVPR}. We investigate whether they transfer from the unimodal setting in which they were introduced to the frozen vision backbones of VLMs, and how they behave on either side of the multimodal projector. We first formalize the layer-profiling task (Section~\ref{subsec:problem}), then review the two geometric tools we adopt (Section~\ref{subsec:prelim}), and finally describe how we instantiate them as layer-wise visual profiles (Section~\ref{subsec:profiling}).

\subsection{Problem Definition}
\label{subsec:problem}

\noindent\textbf{Background.}
A modern VLM connects a vision tower, such as CLIP~\cite{radford2021learning} or SigLIP~\cite{Zhai_2023_ICCV}, to a large language model (LLM) through a multimodal projector~\cite{zhang2024llava,lin2024video}. The vision tower encodes an input image or video into token-level features at each of its $L$ layers. In standard implementations, features from a single fixed layer, commonly the penultimate layer, are passed to the projector, which maps them into the LLM embedding space for downstream reasoning. Crucially, during multimodal instruction tuning the vision backbone is typically kept frozen while the projector is trained. As a result, the visual features \emph{before} the projector are inherited directly from vision pretraining, whereas the features \emph{after} the projector are additionally reshaped by multimodal training. This asymmetry motivates analyzing layer-wise representations separately on the two sides of the projector.

\noindent\textbf{Task.}
Let the vision tower expose candidate layers $l \in \mathcal{L} = \{1,\ldots,L\}$, and
\begin{equation}
	\mathcal{D}_s = \{(x_i, q_i)\}_{i=1}^{N}
\end{equation}
be a small set of task samples, where $x_i$ is an image or video and $q_i$ is the corresponding question or task prompt. Let $\mathrm{Acc}(l)$ denote the downstream task performance obtained when the VLM reads visual features from layer $l$, and
\begin{equation}
	l_{\mathrm{best}} = \arg\max_{l \in \mathcal{L}} \mathrm{Acc}(l)
\end{equation}
be the oracle layer identified by exhaustive layer-wise inference. Evaluating $\mathrm{Acc}(l)$ for every layer requires a full downstream pass per candidate and is prohibitively expensive as $\mathcal{L}$ and the benchmark grow.

The goal of \emph{layer-wise visual profiling} is to find a training-free score $s(l)$, computed from $\mathcal{D}_s$ alone, \ie, without answer generation, gradient updates, or downstream inference, whose top-ranked layers contain the oracle layer. We call such a score effective if, for a small budget $k$,
\begin{equation}
	l_{\mathrm{best}} \in \mathrm{TopK}_k\big(s(l)\big),
\end{equation}
so that profiling narrows the search from all $L$ layers to a compact candidate set that can then be verified with far fewer full-inference runs. Whether any purely geometric, training-free score satisfies this criterion for VLM vision backbones is not known a priori. We therefore treat several candidate signals derived from representation geometry (introduced in Section~\ref{subsec:prelim}) as competing hypotheses, and our profiling process (Section~\ref{subsec:profiling}) decides which, if any, is a reliable selector.

\subsection{Preliminaries}
\label{subsec:prelim}

We build our profiles on two existing training-free geometric tools. Both operate on a representation matrix whose rows are sample-level embeddings, and neither requires labels or answer generation.

\noindent\textbf{Matrix-based entropy.}
Matrix-based entropy quantifies the effective diversity of a set of representations through the spectrum of their Gram matrix~\cite{skean2023dime,skean2025layer}. Given a representation matrix $Z \in \mathbb{R}^{N \times d}$ with one embedding per sample, we center and normalize each row,
\begin{equation}
	\bar{z}^i = \frac{z^i - \mu}{\|z^i - \mu\|_2 + \epsilon},
	\qquad
	\mu = \frac{1}{N}\sum_{i=1}^{N} z^i,
\end{equation}
and form the trace-normalized Gram matrix
\begin{equation}
	A = \frac{\bar{Z}\bar{Z}^\top}{\mathrm{tr}(\bar{Z}\bar{Z}^\top)},
\end{equation}
where $\bar{Z}$ stacks the normalized rows $\bar{z}^i$. With $\{\lambda_j(A)\}$ the eigenvalues of $A$, the entropy and its max-entropy-normalized form are
\begin{equation}
	\mathcal{H}(Z) = -\sum_j \lambda_j(A)\log \lambda_j(A),
	\qquad
	\widehat{\mathcal{H}}(Z) = \frac{\mathcal{H}(Z)}{\log \min(N,d)} .
\end{equation}
A higher $\widehat{\mathcal{H}}(Z)$ indicates that the sample embeddings spread across many directions, while a lower value indicates concentration in a few dominant directions. Prior work established this measure for unimodal language and vision representations; we adopt it unchanged and apply it to visual layers, referring to the resulting quantity as visual dataset entropy (VDE) in Section~\ref{subsec:profiling}.

\noindent\textbf{Gromov--Wasserstein distance.}
Matrix entropy characterizes a single representation space but says nothing about whether two spaces are structurally compatible. Gromov--Wasserstein (GW) distance compares the relational geometry of two representation sets without requiring them to share a coordinate system~\cite{peyre2016gromov,Li_2026_CVPR}. Given a source matrix $P$ and a target matrix $Q$, each with $N$ rows, we form intra-set pairwise distance matrices
\begin{equation}
	D_P(i,j)=d(p_i,p_j),
	\qquad
	D_Q(i,j)=d(q_i,q_j),
\end{equation}
where $d(\cdot,\cdot)$ is the angular cosine distance. Following~\cite{Li_2026_CVPR}, we match the scales of the two spaces by median normalization,
\begin{equation}
	\widetilde{D}_P = \frac{\mathrm{median}(D_Q)}{\mathrm{median}(D_P)}\, D_P,
\end{equation}
and define the GW distance as
\begin{equation}
	\mathrm{GW}(P,Q)
	=
	\min_{\pi \in \Pi(a,b)}
	\sum_{i,j,k,m}
	\left| \widetilde{D}_P(i,j) - D_Q(k,m) \right|
	\pi_{ik}\pi_{jm},
\end{equation}
where $\pi$ is the transport plan and $a,b$ are uniform marginals. A lower $\mathrm{GW}(P,Q)$ indicates that the two sets share more similar internal geometry. This tool was introduced for encoder-level VLM model selection~\cite{Li_2026_CVPR}; we repurpose it as a layer-level cross-modal signal in Section~\ref{subsec:profiling}, and assess its usefulness for layer selection empirically alongside VDE.

\subsection{Layer-wise Visual Representation Profiling}
\label{subsec:profiling}
\label{subsec:gw}

We now instantiate the two tools above on the layer-wise representations of a frozen VLM to obtain a training-free profile for every candidate layer.

\noindent\textbf{Layer-wise visual representations.}
Let $f_l$ map an input to the hidden tokens of the $l$-th vision layer:
\begin{equation}
	H_l^i = f_l(x_i) \in \mathbb{R}^{T_i \times d_v},
\end{equation}
where $T_i$ is the number of visual tokens (including tokens from all sampled frames for video) and $d_v$ is the vision hidden dimension. We reduce each sample to a pre-projector embedding by mean pooling and stack all samples:
\begin{equation}
	z_{l,\mathrm{pre}}^i = \frac{1}{T_i}\sum_{t=1}^{T_i} H_{l,t}^i,
	\qquad
	Z_{l,\mathrm{pre}} = [z_{l,\mathrm{pre}}^1,\ldots,z_{l,\mathrm{pre}}^N]^\top \in \mathbb{R}^{N \times d_v}.
\end{equation}
The multimodal projector $g(\cdot)$ maps visual tokens into the language hidden space; for the MLP projectors used in our models, $g(H_l^i) = W_2\,\sigma(W_1 H_l^i)$ with nonlinearity $\sigma$. The corresponding post-projector representation is
\begin{equation}
	z_{l,\mathrm{post}}^i = \frac{1}{T_i}\sum_{t=1}^{T_i} g(H_l^i)_t,
	\qquad
	Z_{l,\mathrm{post}} = [z_{l,\mathrm{post}}^1,\ldots,z_{l,\mathrm{post}}^N]^\top .
\end{equation}
Comparing profiles computed from $Z_{l,\mathrm{pre}}$ and $Z_{l,\mathrm{post}}$ lets us separate the intrinsic diversity of a vision layer from the diversity actually delivered to the LLM after projection.

\noindent\textbf{Question-side language representations.}
For the GW signal, we also extract a language-side embedding for each sample, without answer generation. We feed the question, or the question with options when available, into the frozen LLM and take the last valid token representation from the penultimate layer:
\begin{equation}
	u_i = h_{-2,\mathrm{last}}(q_i),
	\qquad
	U = [u_1,\ldots,u_N]^\top .
\end{equation}

\noindent\textbf{Layer-wise profiles.}
Applying matrix entropy to the visual representation matrices yields the pre- and post-projector visual dataset entropy,
\begin{equation}
	\mathrm{VDE}_{\mathrm{pre}}(l) = \widehat{\mathcal{H}}(Z_{l,\mathrm{pre}}),
	\qquad
	\mathrm{VDE}_{\mathrm{post}}(l) = \widehat{\mathcal{H}}(Z_{l,\mathrm{post}}),
\end{equation}
where $\mathrm{VDE}_{\mathrm{pre}}$ measures the intrinsic diversity of a vision layer and $\mathrm{VDE}_{\mathrm{post}}$ characterizes how the projector reshapes it. Applying the GW distance between each visual representation matrix and the question-side matrix $U$ yields the cross-modal alignment signals
\begin{equation}
	\mathrm{GW}_{\mathrm{pre}}(l) = \mathrm{GW}(Z_{l,\mathrm{pre}}, U),
	\qquad
	\mathrm{GW}_{\mathrm{post}}(l) = \mathrm{GW}(Z_{l,\mathrm{post}}, U),
\end{equation}
where a lower value indicates that the visual sample geometry is more compatible with the question-side language geometry. The full training-free profile of layer $l$ is
\begin{equation}
	\mathcal{P}(l)
	=
	\{
	\mathrm{VDE}_{\mathrm{pre}}(l),
	\mathrm{VDE}_{\mathrm{post}}(l),
	\mathrm{GW}_{\mathrm{pre}}(l),
	\mathrm{GW}_{\mathrm{post}}(l)
	\}.
\end{equation}

\noindent\textbf{Profiling protocol.}
We treat the entries of $\mathcal{P}(l)$ as candidate layer-selection signals, and evaluate them without any learned or task-specific combination of the four quantities. For a given signal, we rank all candidate layers by that signal, and take the Top-3 as the predicted candidates. Specifically, we rank in descending order for $\mathrm{VDE}$ and ascending order for $\mathrm{GW}$, so that the most favorable layers come first. Following the task definition in Section~\ref{subsec:problem}, we count a hit when the oracle layer $l_{\mathrm{best}}$ from exhaustive layer-wise inference falls within these Top-3, and report per-signal hit rates. This protocol introduces no tunable scoring parameters and directly measures, for each signal, whether the geometry of a small sample set can narrow the visual-layer search space. 

\section{Experiments}

\subsection{Experimental Setup}
\label{subsec:Experimental Setup}

\noindent\textbf{Models and datasets.}
Experiments are conducted on two representative VLMs with different vision backbones: LLaVA-Video-7B-Qwen2~\cite{zhang2024llava} with a 27-layer SigLIP vision tower, and Video-LLaVA-7B-hf~\cite{lin2024video} with a 24-layer CLIP vision tower. Both models connect the vision encoder to the language model through a multimodal projector and use the penultimate visual layer ($-2$) as the default feature layer. All candidate visual layers are evaluated using the corresponding frozen pretrained checkpoints.

For image-based evaluation, we use ScienceQA~\cite{lu2022learn}, POPE~\cite{li2023evaluating}, MMStar~\cite{chen2024we}, and CV-Bench~\cite{tong2024cambrian}. ScienceQA is evaluated on the closed-choice validation subset, POPE on the adversarial split, and CV-Bench on both its 2D and 3D subsets. For video-based evaluation, we use HD-EPIC~\cite{perrett2025hd}, including Action Recognition and Gaze Estimation. Gaze Estimation is evaluated on the full benchmark, while Action Recognition is evaluated on a randomly sampled subset of 500 examples for efficiency. 
A summary of the evaluated datasets is provided in Table~\ref{tab:datasets}.

\begin{table}[t]
\centering
\caption{Datasets used in our experiments.}
\label{tab:datasets}
\begin{tabular}{l l r @{\hspace{1.2em}} l}
\toprule
Domain & Dataset & Samples & Task Type \\
\midrule
Image & ScienceQA & 2,004 & Multi-choice VQA \\
Image & POPE (Adv.) & 3,000 & Hallucination Detection \\
Image & CV-Bench 2D & 1,438 & Count \& Relation Reasoning \\
Image & CV-Bench 3D & 1,200 & Depth \& Distance Reasoning \\
Image & MMStar & 1,500 & General Multimodal Reasoning \\
\midrule
Video & HD-EPIC (Action) & 500 & Egocentric Action Recognition \\
Video & HD-EPIC (Gaze) & 1,000 & Egocentric Gaze Estimation \\
\bottomrule
\end{tabular}
\end{table}

\noindent\textbf{Profiling setup.}
Each dataset is profiled using 100 randomly sampled task examples over five random seeds. 
For every candidate visual layer, we compute $\mathrm{VDE}_{\mathrm{pre}}$ and $\mathrm{VDE}_{\mathrm{post}}$ from pre- and post-projector visual representations, respectively. 
For cross-modal analysis, question-side representations are extracted following Section~\ref{subsec:gw}, and $\mathrm{GW}(Z_{\mathrm{pre}},U)$ and $\mathrm{GW}(Z_{\mathrm{post}},U)$ are computed to measure visual-language structural compatibility. 
The profiling stage is training-free and does not require answer generation or downstream inference, making it substantially cheaper than exhaustive layer-wise evaluation.

\noindent\textbf{Evaluation metrics.}
Layer-wise profiling signals are compared with exhaustive layer-wise downstream accuracy. For VDE, we compare $\mathrm{VDE}_{\mathrm{pre}}$ and $\mathrm{VDE}_{\mathrm{post}}$ with exhaustive layer-wise accuracy to examine whether high-entropy regions correspond to high-performing visual layers, and whether the performance-related trends before projection are preserved after the multimodal projector.

For cross-modal analysis, we compute $\mathrm{GW}(Z_{\mathrm{pre}}, U)$ and $\mathrm{GW}(Z_{\mathrm{post}}, U)$, where $U$ denotes question-side language representations extracted from the last valid token of the penultimate language-model layer. Lower GW indicates stronger structural compatibility between visual and language sample geometries, allowing us to assess whether cross-modal alignment is associated with stronger visual layers.

Quantitatively, we report the Pearson correlation coefficient ($r$), the Spearman rank correlation coefficient ($\rho$), and the coefficient of determination ($R^2$) between each layer-wise profile and downstream accuracy. We also report Top-3 hit rates, where a hit is counted if the best-performing layer from exhaustive inference is included in the Top-3 layers ranked by a profiling signal. 

The analysis is first conducted on ScienceQA, which supports both VDE and GW analysis, and then extended to additional image and video benchmarks.

\subsection{Layer-wise Profiling on ScienceQA}

\begin{figure}[t]
    \centering
    \includegraphics[width=\textwidth]{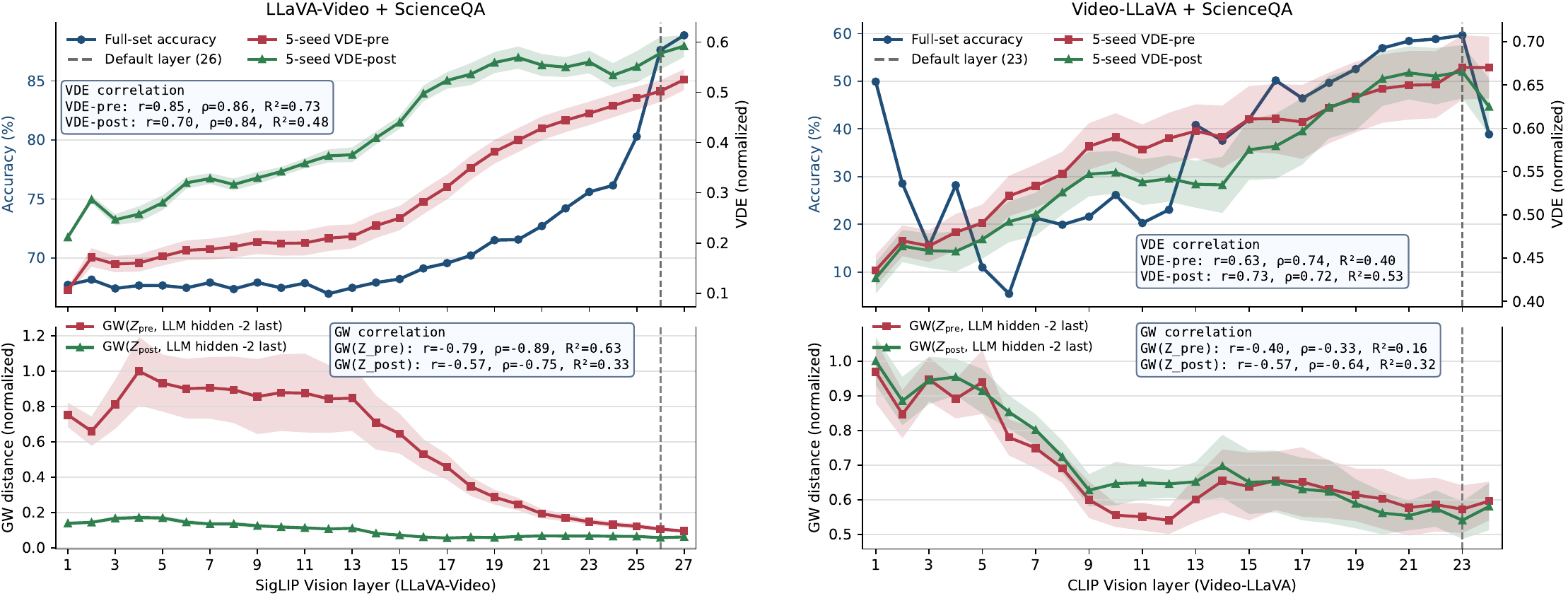}
    \caption{
Layer-wise visual representation profiling on ScienceQA.
VDE is estimated from 100 sampled examples over five seeds and compared with exhaustive layer-wise downstream accuracy. GW distances measure the structural compatibility between visual and question-side language representations.
}
    \label{fig:combined_scienceqa}
\end{figure}

ScienceQA provides semantically rich image-question pairs with multiple-choice options, making it particularly suitable for comparing VDE and GW with downstream accuracy within the same setting. Figure~\ref{fig:combined_scienceqa} reports their layer-wise profiles for both LLaVA-Video and Video-LLaVA.

VDE closely tracks downstream accuracy in both models. For LLaVA-Video, both $\mathrm{VDE}_{\mathrm{pre}}$ and $\mathrm{VDE}_{\mathrm{post}}$ peak at Layer~27, which is also the best-performing layer and differs from the default Layer~26. For Video-LLaVA, the VDE peaks coincide with Layer~23, which is both the default and best-performing layer. Despite using only 100 profiling samples, $\mathrm{VDE}_{\mathrm{pre}}$ shows strong correlations with accuracy ($r=0.85$, $\rho=0.86$, $R^2=0.73$ for LLaVA-Video; $r=0.63$, $\rho=0.74$, $R^2=0.40$ for Video-LLaVA), indicating that sample-level representation diversity captures meaningful layer-wise performance trends before answer generation.

The projector reshapes, but does not erase, the entropy signal. While $\mathrm{VDE}_{\mathrm{pre}}$ follows a smoother and nearly monotonic depth-wise trend as predicted in~\cite{skean2025layer}, $\mathrm{VDE}_{\mathrm{post}}$ exhibits a less regular profile after multimodal projection. Nevertheless, its high-value region remains aligned with the high-performing layers identified by $\mathrm{VDE}_{\mathrm{pre}}$ and downstream accuracy. This suggests that the projector changes the detailed geometry of visual representations while preserving the performance-relevant layer-wise signal.

GW provides a complementary alignment diagnostic rather than a reliable selector. Pre-projector GW generally decreases with depth, suggesting stronger structural compatibility between deeper visual representations and question-side language representations. After projection, GW becomes uniformly low across layers, reducing the contrast needed to localize the best visual layer. Thus, GW helps interpret visual-language alignment, whereas VDE provides the more useful signal for candidate-layer profiling.

\subsection{Generalization Across Image Benchmarks}

\begin{figure*}[t]
    \centering
    \includegraphics[width=\textwidth]{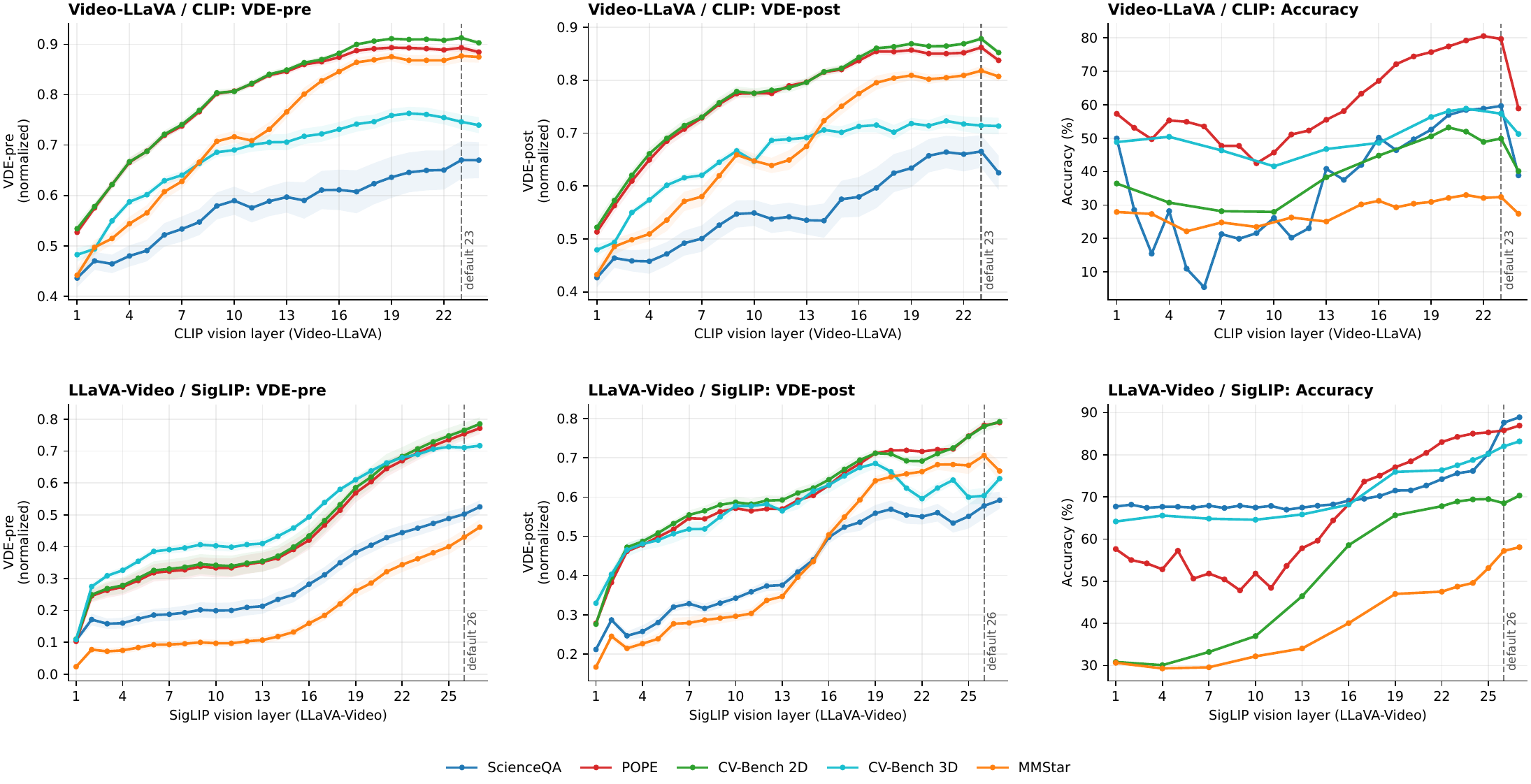}
    \caption{
    Layer-wise $\mathrm{VDE}_{\mathrm{pre}}$, $\mathrm{VDE}_{\mathrm{post}}$, and downstream accuracy across five image benchmarks for Video-LLaVA (CLIP) and LLaVA-Video (SigLIP). Curves are averaged over five random seeds using 100 samples per seed. The dashed vertical line indicates the default visual layer used by each model.
    }
    \label{fig:image_generalization}
\end{figure*}

To evaluate the generality of this profiling behavior, we extend the analysis to four additional image benchmarks covering hallucination detection, spatial reasoning, and general multimodal reasoning. Following the profiling setup in Section~\ref{subsec:Experimental Setup}, all image benchmarks use 100 randomly sampled examples over five random seeds. Figure~\ref{fig:image_generalization} presents the layer-wise $\mathrm{VDE}_{\mathrm{pre}}$, $\mathrm{VDE}_{\mathrm{post}}$, and downstream accuracy profiles across all image benchmarks.

For Video-LLaVA with the CLIP vision tower, $\mathrm{VDE}_{\mathrm{pre}}$ increases from shallow to deeper layers and forms a broad plateau around Layers~18--23, followed by a mild decline at the final layer. Although $\mathrm{VDE}_{\mathrm{post}}$ is less regular after projection, its high-value region remains concentrated in the same upper-layer interval. The downstream accuracy curves show a similar pattern: performance becomes relatively stable from around Layer~18, and the strongest layers are located within the high-VDE region. The final-layer drop in both VDE and accuracy further suggests that representation diversity captures layer-wise performance degradation.

For LLaVA-Video with the SigLIP vision tower, the high-VDE and high-accuracy regions shift toward the deepest layers. $\mathrm{VDE}_{\mathrm{pre}}$ increases progressively with depth, while $\mathrm{VDE}_{\mathrm{post}}$ again has a less monotonic profile but peaks in the same late-layer region. The downstream accuracy curves show a stable high-performance region around Layers~22--27, with Layer~27 consistently outperforming the default Layer~26. This indicates that the preferred representation depth depends on the visual backbone, while the association between high VDE and strong downstream performance remains consistent.

Overall, these image-based results show that the relationship between VDE and downstream performance generalizes beyond ScienceQA and remains consistent across different task types and visual backbones.
Across different image tasks and visual backbones, VDE identifies compact high-performing layer regions, while the post-projector profiles show that the multimodal projector reshapes but does not remove the performance-related entropy signal.

\subsection{Extension to Video Tasks}
\label{subsec:video_extension}

We further examine whether the same profiling behavior extends to video-based VLM tasks. Because video inputs contain multiple frames and are more expensive to profile, we use reduced-budget validation protocols. For HD-EPIC Action Recognition, we perform a full-layer scan using 50 sampled examples with one seed, and additionally compute GW because the answer options provide rich video-text descriptions. For HD-EPIC Gaze Estimation, we use 100 sampled examples over five seeds but restrict the scan to mid-to-high visual layers, following the image results where strong layers concentrate in the upper part of the vision tower. Video-LLaVA uses 8 input frames, while LLaVA-Video uses 16 frames.

\begin{figure*}[t]
    \centering
    \includegraphics[width=\textwidth]{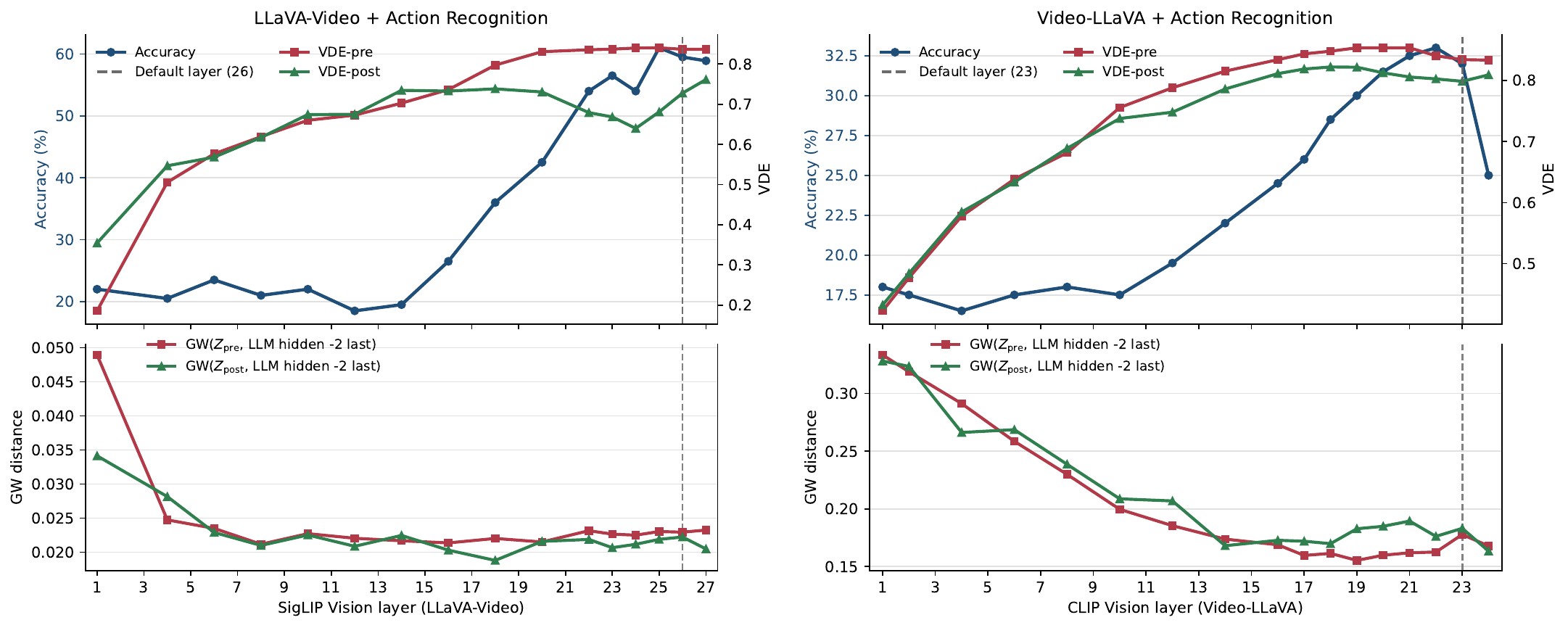}
    \caption{
    Layer-wise profiling results on HD-EPIC Action Recognition. We perform a full-layer scan using 50 sampled examples with one random seed. The figure compares $\mathrm{VDE}_{\mathrm{pre}}$, $\mathrm{VDE}_{\mathrm{post}}$, downstream accuracy, and GW-based cross-modal alignment across visual layers for the evaluated video VLMs.
    }
    \label{fig:action_recognition}
\end{figure*}

Figure~\ref{fig:action_recognition} shows that the image-based findings transfer to video action recognition under the lightweight 50-sample setting. For both video VLMs, $\mathrm{VDE}_{\mathrm{pre}}$ generally increases toward the upper part of the vision tower, while $\mathrm{VDE}_{\mathrm{post}}$ is less regular but still highlights the late-layer region. The downstream accuracy curves follow the same coarse structure: shallow and middle layers are substantially weaker, whereas the strongest layers appear in the upper-layer regions indicated by VDE. For LLaVA-Video, the best-performing layer shifts to around Layer~25, while the default Layer~26 remains competitive. For Video-LLaVA, the strongest layers also lie within the upper-layer plateau, followed by a clear degradation at the final layer. The GW profiles decrease rapidly with depth and stay low in the high-performing region, again supporting their role as an alignment diagnostic rather than a precise selector.

\begin{figure*}[t]
    \centering
    \includegraphics[width=\textwidth]{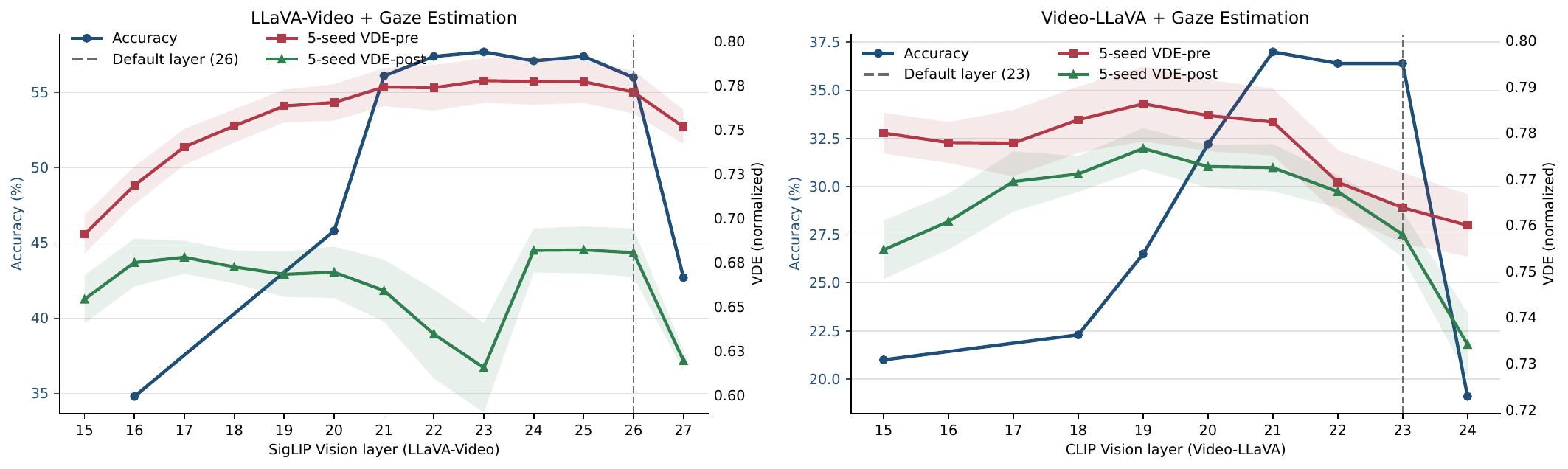}
    \caption{
    Layer-wise profiling results on HD-EPIC Gaze Estimation. We use five random seeds with 100 sampled examples and restrict the scan to the mid-to-high visual layers. The figure compares $\mathrm{VDE}_{\mathrm{pre}}$, $\mathrm{VDE}_{\mathrm{post}}$, and downstream accuracy across visual layers for the evaluated video VLMs.
    }
    \label{fig:gaze_estimation}
    \vspace{-3mm}
\end{figure*}

Figure~\ref{fig:gaze_estimation} further evaluates VDE on HD-EPIC Gaze Estimation under the five-seed setting. For LLaVA-Video, $\mathrm{VDE}_{\mathrm{pre}}$ closely follows downstream accuracy, forming a high-value region around Layers~21--26 that overlaps with the high-performance region. Both VDE and accuracy drop at the final layer, indicating that the entropy profile captures this layer-wise degradation. For Video-LLaVA, the correspondence is weaker but still informative: high VDE concentrates in the mid-to-upper CLIP layers, while both VDE and accuracy decrease at the final layer. Together, the two video tasks show that VDE-based profiling remains practical under reduced search budgets, providing reliable high-performing regions for LLaVA-Video and useful region-level guidance for Video-LLaVA.

\subsection{Top-3 Candidate Layer Retrieval}
\label{subsec:top3_retrieval}

Finally, we summarize the practical utility of VDE-based profiling as a candidate-layer retrieval experiment. For each model-task pair, we rank all evaluated visual layers according to either $\mathrm{VDE}_{\mathrm{pre}}$ or $\mathrm{VDE}_{\mathrm{post}}$ and select the Top-3 layers as candidate layers. We then compare these candidates with the best-performing layer obtained from exhaustive layer-wise downstream inference. A hit is counted if the best-performing layer is included in the Top-3 VDE-ranked candidates.
We also report the downstream accuracy of the Top-1 VDE-selected layer against the oracle best accuracy.

\begin{table*}[htbp]
\centering
\caption{
Top-3 candidate layer retrieval. Hits denote whether the oracle best layer is covered by the Top-3 layers ranked by $\mathrm{VDE}_{\mathrm{pre}}$ or $\mathrm{VDE}_{\mathrm{post}}$. Pre/Post Top-1 vs. Best Acc. compares Top-1 VDE-selected accuracy with oracle best accuracy. Bold entries mark covered oracle layers or exact Top-1 matches.
}
\label{tab:top3_retrieval}
\resizebox{\textwidth}{!}{
\begin{tabular}{c l @{\hspace{1.0em}} c @{\hspace{1.0em}} c @{\hspace{1.0em}} c @{\hspace{1.0em}} c @{\hspace{1.0em}} c @{\hspace{1.0em}} c}
\toprule
Model & Task & VDE-pre Top-3 & VDE-post Top-3 & Pre Top-1/Best Acc. (\%) & Post Top-1/Best Acc. (\%) & Best Layer & Top-3 Hit: Pre/post \\
\midrule
\modelcell{7}{LLaVA-Video}
& ScienceQA        & \textbf{27}, 26, 25 & \textbf{27}, 26, 20 & \textbf{88.90 / 88.90} & \textbf{88.90 / 88.90} & 27 & \cmark / \cmark \\
& POPE-Adv.        & \textbf{27}, 26, 25 & \textbf{27}, 26, 25 & \textbf{87.00 / 87.00} & \textbf{87.00 / 87.00} & 27 & \cmark / \cmark \\
& CV-Bench 2D      & \textbf{27}, 26, 25 & \textbf{27}, 26, 25 & \textbf{70.20 / 70.20} & \textbf{70.20 / 70.20} & 27 & \cmark / \cmark \\
& CV-Bench 3D      & \textbf{27}, 25, 26 & 19, 18, 20 & \textbf{83.17 / 83.17} & 75.92 / 83.17 & 27 & \cmark / \xmark \\
& MMStar           & \textbf{27}, 26, 25 & 26, 24, 23 & \textbf{58.07 / 58.07} & 57.20 / 58.07 & 27 & \cmark / \xmark \\
& HD-EPIC Action   & \textbf{25}, 24, 23 & 27, 18, 14 & \textbf{61.00 / 61.00} & 58.90 / 61.00 & 25 & \cmark / \xmark \\
& HD-EPIC Gaze     & \textbf{23}, 24, 25 & 25, 24, 26 & \textbf{57.70 / 57.70} & 57.40 / 57.70 & 23 & \cmark / \xmark \\
\midrule
\modelcell{7}{Video-LLaVA}
& ScienceQA        & \textbf{23}, 24, 22 & \textbf{23}, 21, 22 & \textbf{59.63 / 59.63} & \textbf{59.63 / 59.63} & 23 & \cmark / \cmark \\
& POPE-Adv.        & 19, 23, 20 & 23, 19, 17 & 76.80 / 80.53 & 79.80 / 80.53 & 22 & \xmark / \xmark \\
& CV-Bench 2D      & 23, 19, 21 & 23, 22, 19 & 49.86 / 53.20 & 49.86 / 53.20 & 20 & \xmark / \xmark \\
& CV-Bench 3D      & 20, \textbf{21}, 19 & \textbf{21}, 19, 22 & 58.08 / 58.83 & \textbf{58.83 / 58.83} & 21 & \cmark / \cmark \\
& MMStar           & 23, 19, 24 & 23, 19, 22 & 32.40 / 33.00 & 32.40 / 33.00 & 21 & \xmark / \xmark \\
& HD-EPIC Action   & 20, 19, 21 & 19, 18, 17 & 31.50 / 33.00 & 30.00 / 33.00 & 22 & \xmark / \xmark \\
& HD-EPIC Gaze     & 19, 20, 18 & 19, 20, \textbf{21} & 26.50 / 37.00 & 26.50 / 37.00 & 21 & \xmark / \cmark \\
\bottomrule
\end{tabular}
}
\vspace{-4mm}
\end{table*}

Table~\ref{tab:top3_retrieval} further summarizes the candidate-layer retrieval results. For LLaVA-Video with the SigLIP visual tower, the Top-3 candidates ranked by $\mathrm{VDE}_{\mathrm{pre}}$ consistently cover the oracle best layer across both image and video tasks. This suggests that the VDE-ranked layers provide a useful approximation to the true high-performing layer region. In particular, the same behavior holds not only on image benchmarks, where the best layer is often the final visual layer, but also on the video Action Recognition task, where the best layer shifts to Layer~25 while still remaining within the high-VDE late-layer region.

For Video-LLaVA with the CLIP visual tower, strict Top-3 retrieval is less stable. 
The model often exhibits a broad upper-layer plateau, where adjacent layers have similar VDE values and comparable downstream accuracy. As a result, VDE is less reliable as an exact Top-1 or strict Top-3 selector, but still provides useful region-level guidance in several tasks by concentrating candidates around the stronger upper-layer region. 

Overall, $\mathrm{VDE}_{\mathrm{pre}}$ is the strongest practical signal among the tested profiles. It reliably retrieves high-performing layers for the SigLIP-based LLaVA-Video model and offers region-level guidance for the CLIP-based Video-LLaVA model. GW is better interpreted as a diagnostic of cross-modal alignment rather than a dependable stand-alone layer selector. These results support VDE, especially $\mathrm{VDE}_{\mathrm{pre}}$, as a lightweight training-free policy for narrowing visual-layer search.

\paragraph{Limitations.}

First, the experiments focus on two representative VLMs, LLaVA-Video with SigLIP and Video-LLaVA with CLIP. Although the benchmarks cover both image and video tasks, further validation on broader model families, larger vision towers, and different projector designs is needed.
Second, VDE is a profiling signal rather than a guaranteed exact layer selector. It reliably retrieves high-performing layers for the SigLIP-based model, but is less stable as a strict Top-3 or Top-1 selector for the CLIP-based model. In such cases, VDE is better viewed as a region-level policy that narrows the search space and should be followed by limited downstream verification.
Third, GW distance is only diagnostic in our framework. 
It provides evidence of visual-language alignment across depth, but does not by itself offer a dependable layer-selection rule.

\section{Conclusion}

This paper studies visual-layer choice in LLaVA-style VLMs from a training-free representation profiling perspective. Instead of treating the late visual layer used by these models as a fixed architectural convention, we analyze how downstream performance changes across vision depth and how this behavior relates to layer-wise representation geometry. Across image and video benchmarks, the results show that the default layer is not always optimal, and that high-performing layers often form compact regions that can be identified before answer generation. Using Visual Dataset Entropy, we find that sample-level visual diversity provides a simple and effective signal for profiling candidate layers. The pre-projector entropy profile is especially reliable for the SigLIP-based LLaVA-Video model, while also offering useful region-level guidance for the CLIP-based Video-LLaVA model. The post-projector profile further shows that the multimodal projector reshapes visual geometry but does not fully remove performance-related layer-wise trends. In contrast, Gromov-Wasserstein distance provides useful evidence of visual-language alignment across depth, but is better interpreted as a diagnostic rather than a standalone selector.

Overall, these findings suggest that representation geometry can provide insightful analysis and useful guidance for narrowing the visual-layer search space, reducing reliance on exhaustive inference. However, generalizability of entropy-based profiling across more diverse visual backbones and to different families of vision-language models remains to be studied more systematically.


\section*{Acknowledgements}
This work has been supported by the 
Centre for Spatial Intelligence (RCSI) at University
of Bath,  the European Union under grant agreement no. 101136006-XTREME,
the European Innovation Council under grant agreement no. 1012575-36-CEREBRIS,
the MWK of Lower Saxony within Hybrint (VWZN4219) and LCIS (VWZN4704), the DFG under Germany’s Excellence Strategy within the Cluster of Excellence PhoenixD (EXC2122) and Quantum Frontiers (EXC2123), 
and the German Research Foundation (DFG) as part of the Research Training Group i.c.sens (GRK2159).

\bibliographystyle{splncs04}
\bibliography{main}
\end{document}